\documentclass[conference]{IEEEtran}
\IEEEoverridecommandlockouts
\usepackage{cite}
\usepackage{amsmath,amssymb,amsfonts}
\usepackage{algorithmic}
\usepackage{graphicx}
\usepackage{hyperref}
\usepackage{float}
\usepackage{textcomp}
\usepackage{xcolor}
\usepackage{booktabs}
\usepackage{multirow}
\usepackage{tabularx}
\usepackage[addedmarkup=colored,commentmarkup=todo]{changes}
\definechangesauthor[name={Chidike},color=green]{CV}
\usepackage{xcolor}
\hypersetup{pdfstartview={XYZ null null 1.00}}

\newcommand{\add}[1]{\textcolor{blue}{#1}}

\def\BibTeX{{\rm B\kern-.05em{\sc i\kern-.025em b}\kern-.08em
 T\kern-.1667em\lower.7ex\hbox{E}\kern-.125emX}}

\begin{document}

\title{Privacy-Preserving ECG-HRV Emotion Classification for Mood Swing Detection using Federated Learning on Constrained Edge Devices}

\author{\centering
\IEEEauthorblockN{Chidike Vincent,~Muhammad Aslam Jarwar,~Najam Ul Hasan}
\IEEEauthorblockN{Laurence Hirsch,~Jonathan Zasada-James,~Hem Raj Dadral}
\IEEEauthorblockA{School of Computing and Digital Technologies, Sheffield Hallam University, Sheffield, United Kingdom\\
\{v.vincent, a.jarwar, n.ul-hasan, l.hirsch
j.zasada-james, hemraj.r.dadral\}@shu.ac.uk}
}

\maketitle

\begin{abstract}
This paper presents a privacy-preserving mood monitoring  framework that classifies emotional states from ECG-derived heart rate variability (HRV) features. The proposed framework is evaluated on three 
 datasets comprising 74 participants  and  validated  on an  unseen individual using a  wearable in a pilot study. Leave-One-Participant-Out  cross-validation LOPO-CV, which evaluates the model  on each participant without exposing their data during training, serves as the primary generalization benchmark. A  multi-phase experimental design examines how temporal windowing, HRV feature selection, and class equalisation affects four-class valence-arousal and binary valence classification. Class imbalance is identified as the dominant suppressor of low-valence; and equalisation improves minority class recall by  48.5 \%  on the DREAMER dataset. Binary valence classification achieves  53.17\%  LOPO-CV accuracy using standard HRV features. 
  Under identical conditions, Logistic Regression  outperforms the other five classifiers and exceeding comparable HRV-based baselines.  Pilot validation using a Polar H10 wearable  achieves 75\% binary valence accuracy on an  unseen individual without  calibrations, including 100\% precision for negative valence predictions. 
  Despite being limited to a single participant, the finding demonstrates the framework's feasibility for validation on a larger cohort.
  Finally, the proposed framework is deployed on a six-node Raspberry Pi federated learning cluster to evaluate the privacy-preserving distributed training. The results show that federated Logistic Regression achieves accuracy comparable to centralized Logistic Regression across all three datasets, without sharing raw physiological data, thereby preserving user privacy.  These findings establish the proposed framework as an interpretable, and privacy preserving solution  for continuous mood monitoring in real-world clinical and community settings.
\end{abstract}


\section{Introduction}

Mood disorders in young people often go undetected 
due to reliance on self-reporting and scheduled assessments. Millions of young people experience symptoms of depression, anxiety, and mood disorders without a diagnosis. Because individuals rarely present to a clinic that fits traditional diagnostic frameworks, the average delay between symptom onset and treatment stretches to eight to twelve years \cite{who2022mental}. This gap is not because effective treatments are unavailable, but because  tools for detecting mood changes early enough to intervene are still largely absent from everyday life. Depression is the leading global cause of disability, affecting an estimated 322 million people, with particularly high prevalence among adolescents and young adults \cite{who2017depression}.

The human body does not wait for a clinical appointment to signal distress. The autonomic nervous system responds to emotional states continuously, and those responses are measurable through the heart. Heart rate variability (HRV), defined as the variation in time intervals between successive R-peaks in the electrocardiogram (ECG), reflects the dynamic balance between the sympathetic and parasympathetic nervous systems. Emotions can be described along two dimensions: valence, which captures whether a feeling is positive or negative, and arousal, which captures how activated or calm a person feels \cite{russell1980circumplex}. A person feeling joyful scores high on both; someone feeling sad scores low on valence and low on arousal; someone anxious scores high on arousal but low on valence. HRV is an established physiological correlate of both dimensions \cite{camm1996hrv,valenza2014emotions}. Wearable devices such as chest straps and smartwatches are now widely available and capable of recording cardiac electrical activity with sufficient fidelity to extract meaningful physiological information. This research evaluates whether these signals, processed through intelligent privacy-preserving frameworks, can serve as reliable longitudinal markers of an individual's emotional state.

Despite an increased volume of research applying machine learning to ECG-HRV signals, three limitations hinder clinical translation. First, the majority of existing models rely on within-participant (WP) training protocols, which inflate accuracy and do not generalise to unseen individuals \cite{ferdinando2016ecg,katsigiannis2018dreamer}. Second, high-accuracy models rely on laboratory equipment or computationally intensive feature extraction methods incompatible with resource-constrained edge devices \cite{nita2022cnn,wang2023hrv,8190800}. Third, centralising continuous physiological data raises privacy concerns. The proposed system functions as a personal awareness tool rather than a definitive clinical diagnostic instrument. A cross-participant (XP) accuracy of approximately 53\% at inception is sufficient to detect mood trends and support personalisation over time, and the federated architecture  preserves individual privacy by ensuring personal data remains local while improving performance progressively through continuous on-device learning. 
This paper addresses all three limitations through a unified, systematic experimental framework. The main contributions are as follows:

\begin{itemize}
\item A multi-phase windowed HRV classification approach is evaluated across three public affective datasets, DREAMER \cite{katsigiannis2018dreamer}, AMIGOS \cite{miranda2021amigos}, and EmoWear \cite{rahmani2024emowear}, using Leave-One-Participant-Out cross-validation as the primary generalisation benchmark.
\item Class imbalance is identified and characterised as the primary suppressor of low-arousal low-valence detection, a finding with direct clinical relevance as this quadrant corresponds most closely to depressive mood states.
\item The approach is validated on an unseen participant using a Polar H10 consumer chest strap, demonstrating real-world deployability on hardware accessible to end users without laboratory infrastructure.
\item A privacy-preserving federated learning architecture is implemented and deployed on a six-node Raspberry Pi cluster, showing that distributed training on partitioned physiological data consistently match centralised training while keeping all raw signals local to each device.
\end{itemize}

The remainder of this paper is organised as follows. Section \ref{sec:relw} reviews related work. Section \ref{sec:method} presents the methodology. Section \ref{sec:exp_res} reports experimental results. Section \ref{sec:dep} describes real-world deployment and live validation.  Section \ref{sec:per} evaluates performance. Section \ref{sec:dis} discusses findings and limitations. Section \ref{sec:conc} concludes the paper.



\section{Related Work}\label{sec:relw}

\subsection{ECG-HRV Emotion Classification}

Heart rate variability analysis for emotion recognition has been explored since the early work of Valenza et al.~\cite{valenza2014emotions}, who demonstrated that time and frequency domain HRV features extracted from short ECG windows could discriminate between positive and negative emotional states using a point-process modelling framework. The DREAMER dataset, introduced by Katsigiannis and Ramzan
 \cite{katsigiannis2018dreamer}, established a widely used benchmark for multimodal emotion recognition using low-cost, off-the-shelf ECG and EEG devices. Using a Support Vector Machine on within-participant folds, the original DREAMER paper reported 62.49\% valence accuracy, the primary ECG-based reference for this dataset.
 Subsequent work pursued higher accuracy through increasingly complex methods. Nita et al.~\cite{nita2022cnn} applied a seven-layer convolutional neural network with synthetic data augmentation directly to raw ECG waveforms from DREAMER, reporting 95.16\% within-participant valence accuracy. Wang et al.~\cite{wang2023hrv} proposed an HRV Emotion Recognition method combining Emotion Quantification analysis with Amplitude Level Quantisation to extract frequency-band-specific local amplitude patterns, achieving 84.3\% within-participant valence accuracy using Logistic Regression on DREAMER. Ferdinando et al.~\cite{ferdinando2016ecg} applied bivariate empirical mode decomposition spectral features to ECG from the Mahnob-HCI database, achieving 42.6\% valence accuracy using standard HRV features under 10-fold cross-validation and 59.2\% under subject-independent validation on a three-class problem.

Despite significant advances, four critical limitations persist in the current literature. First, within-participant evaluation artificially inflates accuracy and fails to reflect real-world performance on unseen individuals. Second, deep learning on raw ECG waveforms requires GPU computation, making it incompatible with the resource constraints of edge computing devices. Third, novel features such as Amplitude Level Quantisation (ALQ) lack cross-participant validation and likely encode individual cardiac identity rather than shared emotional signal. Finally, no prior study  has conducted validation on a live participant using a consumer wearable, which is a prerequisite for determining real-world system efficacy. This work addresses all four gaps, surpassing the 42.6\% standard HRV baseline reported by Ferdinando et al.~\cite{ferdinando2016ecg} under the stricter Leave-One-Participant-Out (LOPO-CV) protocol and providing live consumer-hardware validation.

\subsection{Federated Learning for Physiological Monitoring}

Federated learning is a distributed machine learning paradigm in which a collaborative model is trained across multiple devices. Each participant trains a local model on their own private data and shares only the resulting model weight updates with a central aggregation server, which combines them into an improved global model. Raw physiological signals never leave the local device. Federated learning was formally introduced by McMahan et al.~\cite{mcmahan2017fedavg} through the Federated Averaging (FedAvg) algorithm. 
ECG-based physiological signals are inherently non-Independent and Identically Distributed (non-IID) across individuals, with each person presenting a distinct cardiac autonomic baseline \cite{camm1996hrv}, meaning centralising data often leads to models that struggle to generalise across diverse individuals. Federated learning resolves this by allowing each node to develop a locally calibrated model before contributing to the global update, thereby improving robustness across heterogeneous participants. 
The Flower framework \cite{beutel2020flower} provides a hardware-agnostic federated implementation applicable to heterogeneous edge devices.
 Prior work has demonstrated federated approaches that consistently match or exceed centralised accuracy for physiological signal classification whilst preserving privacy \cite{frontiers2024federated,bagherzadeh2024federated}. The present study extends this literature with a physical hardware deployment tested across three independent datasets. 

\section{Methodology}\label{sec:method}

\begin{figure*}[htbp]
\centerline{\includegraphics[width=0.95\textwidth]{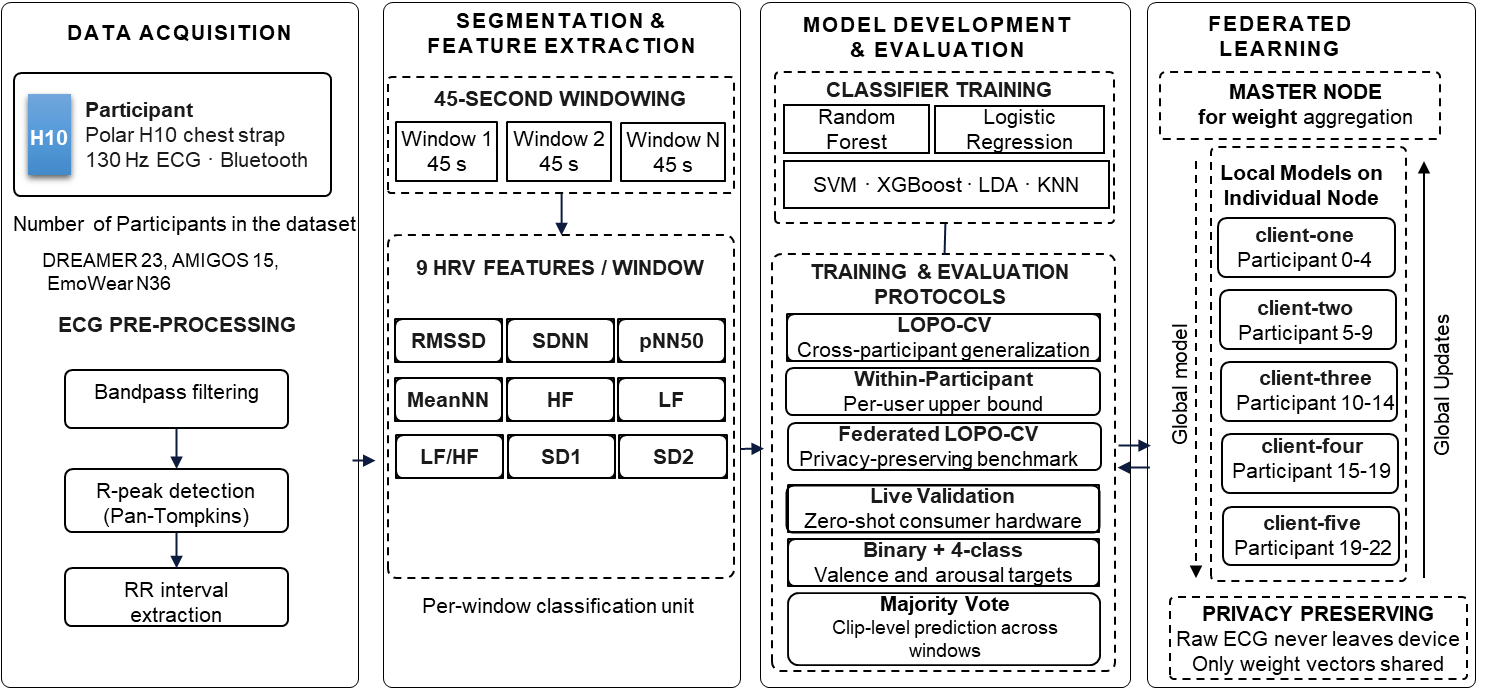}}
\caption{Proposed federated system architecture for ECG-HRV based mood detection.
}
\label{fig:approach}
\end{figure*}

Fig.~\ref{fig:approach} shows the proposed system architecture, covering the full process from raw ECG acquistion through to federated deployment. The models are initially developed using existing datasets, after which ECG data is collected using a Polar H10 chest strap, with signal processing executed locally on the recording device. For both the public datasets and the live recordings, each continuous recording is segmented into 45-second windows from which HRV features are extracted and passed to a classification model. For cross-participant evaluation, participant-level partitioning is performed before model training to ensure that data from the held-out participant are never exposed during training. Finally, in the federated learning stage, only model parameter updates, rather than raw signals, are transmitted to the aggregation server.

\subsection{Temporal Windowing Strategy}

\begin{figure}[htbp]
\centerline{\includegraphics[width=\columnwidth, height=6cm, keepaspectratio,keepaspectratio]{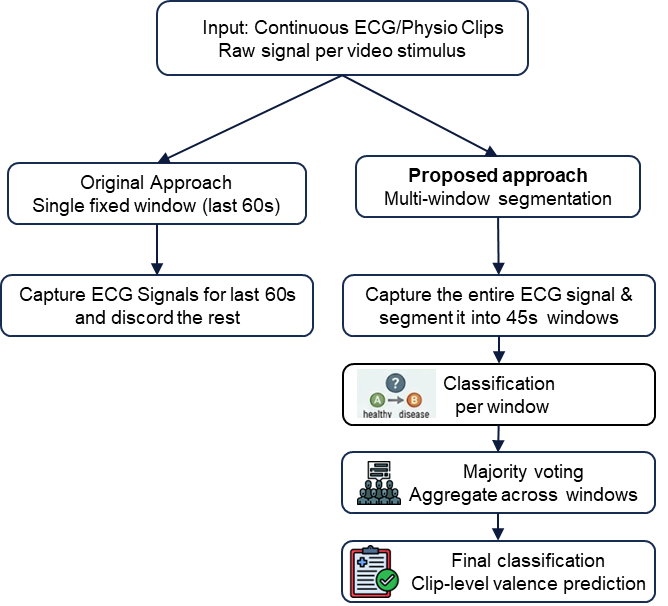}}
\vspace{-5pt}
\caption{Comparison of the original single-window  (left) and the proposed multi-window segmentation approach (right). 
}
\label{fig:windowing}
\end{figure}

The original DREAMER study extracted a single fixed window from the final 60 seconds of each stimulus clip \cite{katsigiannis2018dreamer}. While this provides a baseline, it discards the emotional information present in the rest of the clip and produces only one training sample per clip, limiting data available for cross-participant learning. To overcome this, our approach divides  each clip into non-overlapping 45-second segments. Although, European Society of Cardiology recommends 2 to 5 minutes for short-term HRV spectral analysis \cite{camm1996hrv}, this applies to the cumulative recording duration rather than internal segmentation. Since stimulus clip durations range from 65 to 393 seconds in DREAMER and are 180 seconds in live validation, all recordings satisfy this requirement. Each 45-second window produces an independent prediction, and  the final clip-level classification is determined by the majority voting across all windows. The 45-second duration was selected empirically. 
 To identify the optimal window length, five different window durations were evaluated using the same Random Forest classifier under identical preprocessing, feature extraction,  and evaluation settings as shown in Table~\ref{tab:windows}.
 The 45-second window achieves the highest performance  under both the LOPO-CV and within-participant evaluation protocols and was therefore adopted  for all subsequent phases. The  temporal window selection process is illustrated in Fig.~\ref{fig:windowing}.
\vspace{-0.5em}
\begin{table}[htbp]
\caption{Effect of Window Duration on accuracy (DREAMER Dataset, Four-Class, Random Forest, Balanced Data)}
\label{tab:windows}
\centering
\small
\begin{tabularx}{\columnwidth}{Xcc}
\toprule
\textbf{Duration} & \textbf{LOPO-CV} & \textbf{Within-Participant} \\
\midrule
15 seconds & 33.38\% & 40.05\% \\
30 seconds & 32.41\% & 40.52\% \\
\textbf{45 seconds} & \textbf{38.27\%} & \textbf{43.93\%} \\
60 seconds 
& 34.26\% & 41.02\% \\
90 seconds & 38.19\% & 40.54\% \\
\bottomrule
\end{tabularx}
\end{table}
\vspace{-0.5em}

\subsection{Feature Extraction}
ECG signals were preprocessed using NeuroKit2 \cite{makowski2021neurokit}, including  bandpass filtering, R-peak detection using the Pan-Tompkins algorithm, and RR interval extraction. Each 45-second window produces a single HRV feature vector. Each feature vector includes nine standard time-domain and frequency-domain HRV features: MeanNN, SDNN, RMSSD, pNN50, LF, HF, LF/HF, SD1, and SD2.  Let $RR_i$ denote the interval between the $i^{\mathrm{th}}$ and $(i+1)^{\mathrm{th}}$ consecutive R-peaks in the ECG signal, where $N$ is the total number of RR intervals within each 45-second window. The primary time-domain HRV features are computed as follows:

\begin{equation}
\text{RMSSD} = \sqrt{\frac{1}{N-1}\sum_{i=1}^{N-1}(RR_{i+1} - RR_i)^2}
\end{equation}

\begin{equation}
\text{SDNN} = \sqrt{\frac{1}{N-1}\sum_{i=1}^{N}(RR_i - \overline{RR})^2}
\end{equation}

\begin{equation}
\text{pNN50} = \frac{|\{i : |RR_{i+1} - RR_i| > 50\text{ms}\}|}{N-1} \times 100
\end{equation}

These features are not arbitrary signal statistics. Each captures a distinct aspect of autonomic nervous system function that is physiologically linked to emotional and mood states. Root Mean Square of Successive Differences (RMSSD) and the Percentage of successive Normal-to-Normal intervals exceeding 50\,ms (pNN50) reflect parasympathetic activity, specifically vagal tone, which decreases under psychological stress and negative affect and is chronically reduced in clinical depression \cite{camm1996hrv,pham2021hrv}. SDNN captures overall autonomic regulatory capacity, a global indicator of the body's ability to adapt to environmental demands, which is similarly diminished in prolonged low-mood states. LF power and the LF/HF ratio reflect sympathovagal balance, shifting toward sympathetic dominance during arousal and stress. MeanNN captures average cardiac cycle length, which lengthens during relaxation and shortens during activation. SD1 and SD2 from the Poincar\'{e} plot represent short-term and longer-term beat-to-beat variability respectively, providing complementary views of the dynamic interplay between sympathetic and parasympathetic regulation \cite{pham2021hrv}.




\subsection{Evaluation Protocol}

The proposed framework is evaluated using two evaluation protocols that is LOPO-CV to assess subject-independent generalisation and within-participant evaluation to compare with the previous works. Under the LOPO-CV protocol, the evaluation consists of K folds. In each fold k, participant k is held out of testing while the model is trained on the rest of the participants. After completing all folds, the classification accuracy is computed as the average accuracy across all K folds, as defined in Eq.~(4).  
\vspace{-0.5em}
\begin{equation}
\text{LOPO-CV} = \frac{1}{K}\sum_{k=1}^{N} \text{Accuracy}(M_{-k}, P_k)
\end{equation}

where $M_{-k}$ denotes the model trained without participant $k$ and $P_k$ the held-out test set. 

  For the within-participant evaluation, 5-fold cross-validation was performed independently for each participant, using an 80\%/20\% training/testing split in each fold. The reported results are averaged across all five folds and participants.


\subsection{Classification and Equalisation}
To investigate the effect of class imbalance on four class emotion classification
a Random Forest (RF) classifier with  class\_weight=`balanced' hyperparameter and 200 estimators was used. Balanced weighting adjusts the contribution of each class to training in an inverse proportion to its frequency, providing a baseline correction for imbalance within the classifier itself. To further investigate the impact of class imbalance beyond class weighting, explicit class equalisation was additionally applied in selected phases by capping all classes at the window count of the smallest class. For DREAMER four-class classification, the class distribution includes High Arousal–Low Valence (HALV, 304), High Arousal–High Valence (HAHV 261), Low Arousal–High Valence (LAHV, 274), and Low Arousal–Low Valence (LALV, 109) windows. Consequently, the LALV class contains only 36\% of the windows in the largest class, indicating substantial class imbalance.

To mitigate this imbalance, two balancing strategies were considered: downsampling and synthetic oversampling. Downsampling to the minority class count was selected  instead of synthetic oversampling methods such as SMOTE (Synthetic Minority Oversampling Technique) for two reasons. First, SMOTE generates synthetic HRV feature vectors by interpolating between existing minority class samples. Given the high inter-individual variability of HRV signals, such synthetic samples risk introducing physiologically implausible feature combinations that do not correspond to any real cardiac pattern, as HRV features exhibit high inter-individual variability that makes interpolation between participants unreliable \cite{camm1996hrv,pham2021hrv}. Second, empirical comparison on the DREAMER confirmed that class-weighted RF with equalisation produced better minority class recall than SMOTE augmentation whilst maintaining interpretable training data. Equalisation was not applied to binary valence experiments on the DREAMER dataset because the HIGH and LOW classes were already approximately balanced, containing  544 and 560 windows respectively.  This 1.5\% difference was considered insufficient to  warrant additional class balancing.

\subsection{Federated Learning Framework}


The proposed framework was deployed using a federated learning architecture, with Logistic Regression (LR) adopted as the local classifier on each client node. The federated learning uses FedAvg algorithm to aggregate  the locally trained model's parameters received from each client into a global model.  This aggregation strategy requires that every client's model be representable as a single continuous numerical weight vector. LR satisfies this directly: its parameters are a compact weight matrix of fixed dimensions ($d \times C$, where $d$ is the number of features and $C$ the number of classes) that can be directly averaged  across clients. The LR classifier was configured with regularisation strength $C{=}1.0$ (L2 penalty), the  LBFGS solver, and a maximum of 1000 iterations, consistent across all centralised and federated experiments. Random Forest (RF), whilst used in the centralised four-class experiments, consists of an ensemble of independent decision trees. Decision tree structures are not representable as fixed-length numerical vectors and therefore cannot be averaged by FedAvg in a principled way. Whilst federated variants of RF using alternative aggregation strategies exist \cite{hauschild2022fedrf}, standard FedAvg as used here requires a continuous weight representation. 
The FedAvg algorithm \cite{mcmahan2017fedavg} computes  the global model by averaging locally trained weight vectors 

\begin{equation}
w_{t+1} = \sum_{k=1}^{K} \frac{n_k}{n} w_k^t
\end{equation}

where $w_k^t$ is the weight vector of client $k$ after local training at round $t$, $n_k$ is the number of local training examples, and $n = \sum_k n_k$ is the total across all clients.

 Accordingly,  logistic regression was used throughout the federated learning experiments. 
For the federated  learning experiments, the entire dataset was  partitioned  by participants with each client training a local logistic regression model and returning only the model weight updates for the FedAvg aggregation. 

\subsection{Datasets}

Three publicly available affective computing datasets were selected  to evaluate the proposed framework, namely DREAMER \cite{katsigiannis2018dreamer}, AMIGOS  \cite{miranda2021amigos}, and EmoWear \cite{rahmani2024emowear}. Table~\ref{tab:datasets} summarises their key characteristics. In all three these datasets, emotional states are labelled through self-report on a two-dimensional valence–arousal space after each stimulus clip. For DREAMER (1–5 scale), the midpoint threshold is 3. For AMIGOS and EmoWear (1--9 scale), the threshold is 5. Clips rated exactly at the midpoint on either dimension were labelled ambiguous and excluded from training and evaluation. 

\vspace{-0.5em}
\begin{table}[htbp]
\caption{Dataset Characteristics.  Hz = ECG sampling rate; LALV = Low Arousal Low Valence window count.}
\label{tab:datasets}
\centering
\small
\begin{tabularx}{\columnwidth}{Xcccc}
\toprule
\textbf{Dataset} & \textbf{participants} & \textbf{Hz} & \textbf{Scale} & \textbf{LALV} \\
\midrule
DREAMER \cite{katsigiannis2018dreamer} & 23 & 256 & 1--5 & 109 \\
AMIGOS \cite{miranda2021amigos} & 15 & 128 & 1--9 & 49 \\
EmoWear \cite{rahmani2024emowear} & 36 & 250 & 1--9 & 219 \\
\bottomrule
\end{tabularx}
\end{table}
\vspace{-0.5em}

\section{Experimental Results} \label{sec:exp_res}
This section evaluates our proposed ECG-HRV emotion classification framework. We first establish baseline performance for four-class valence-arousal classification on the DREAMER dataset using a Random Forest (RF) classifier, applying both subject-independent (LOPO-CV) and within-participant protocols. We then examine the impact of class imbalance through equalization. Finally, we validate the framework on the AMIGOS and EmoWear datasets, compare six machine learning classifiers for binary valence classification, and benchmark our results against existing ECG-HRV emotion recognition studies.

\subsection{Baseline and the Class Imbalance Finding}

Following selection of the optimal window  and HRV features, the baseline four-class classification performance was evaluated on the DREAMER dataset. The LOPO-CV accuracy  is 38.27\% while the within-participant accuracy is 43.93\%. 
LALV recall is 6.5\% under LOPO-CV (7.8\% within-participant 
), indicating that the classifier correctly identifies only a small proportion of LALV samples.  This happens due to the underrepresentation of the LALV class (only 12\% of training examples), the classifier was unable to learn an accurate decision boundary for the minority class. 

\begin{figure}[htbp]
\centerline{\includegraphics[width=\columnwidth]{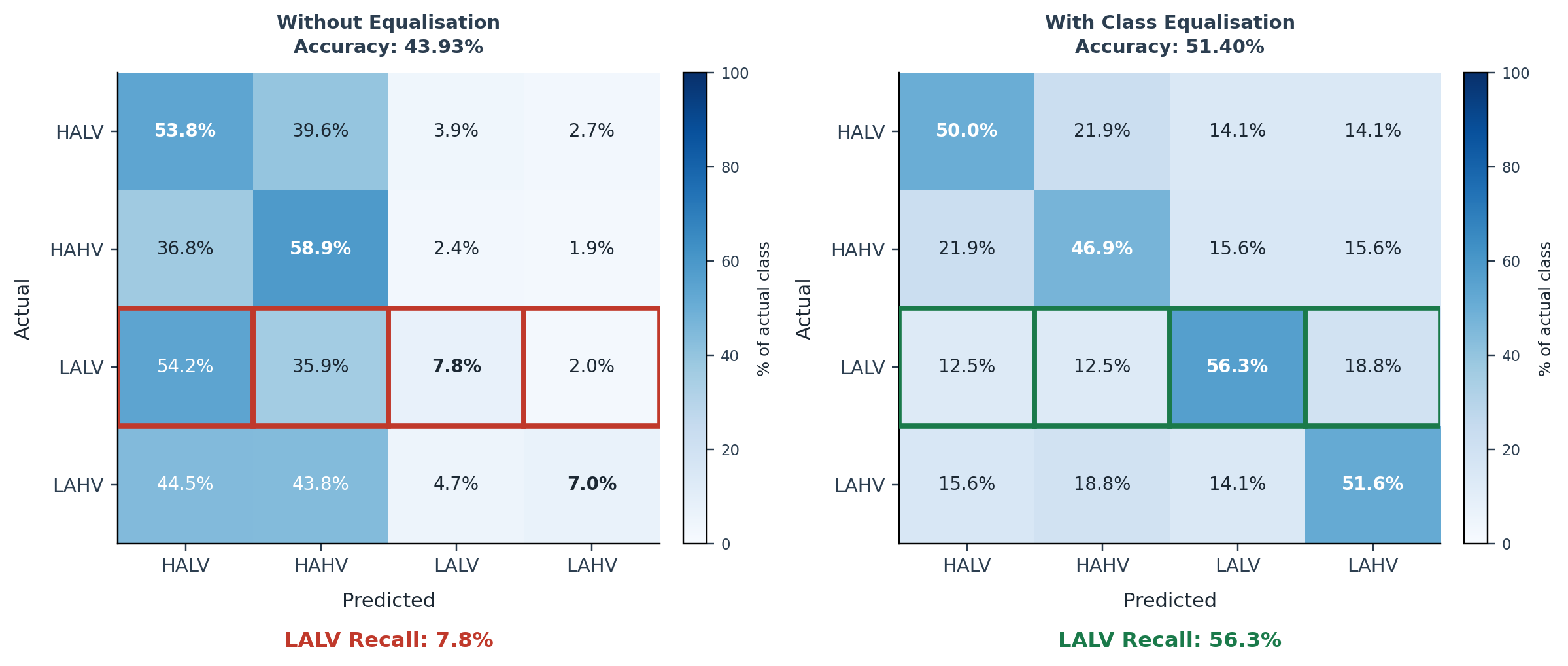}}
\vspace{-5pt}
\caption{Confusion matrices for four-class DREAMER classification.}
\label{fig:confusion}
\end{figure}

To isolate the effect of class imbalance, class equalisation was applied by capping all four classes at 109 windows. All other experimental settings remain the same. 
Table~\ref{tab:equalisation} summarises the quantitative effect of class equalisation on overall accuracy and class-wise recall, while Fig.~\ref{fig:confusion} presents the corresponding row-normalised confusion matrices, illustrating how the class-wise prediction patterns change following equalisation. Together, the results show that class equalisation substantially improves recognition of the minority LALV class while producing more balanced classification performance across all four classes. Overall accuracy increased from 43.93\% to 51.40\%, while LALV recall improved from 7.8\% to 56.3\%, demonstrating substantially improved recognition of the minority class.
\vspace{-0.5em}
\begin{table}[htbp]
\caption{Effect of Class Equalisation (DREAMER, Four-Class, Within-Participant)}
\label{tab:equalisation}
\centering
\scriptsize
\setlength{\tabcolsep}{6pt}
\begin{tabular}{lrrrrr}
\toprule
\textbf{Condition} & \textbf{Acc.} & \textbf{LALV} & \textbf{HALV} & \textbf{HAHV} & \textbf{LAHV} \\
\midrule
No equalisation & 43.93\% & 7.8\% & 53.8\% & 58.9\% & 7.0\% \\
With equalisation & 51.40\% & 56.3\% & 50.0\% & 46.9\% & 51.6\% \\
Change & +7.47pp & +48.5pp & $-$3.8pp & $-$12.0pp & +44.6pp \\
\bottomrule
\end{tabular}
\end{table}
\vspace{-0.5em}

\vspace{-0.5em}

\subsection{Multi-Dataset Evaluation}
Having demonstrated the benefit of class equalisation on the DREAMER dataset, the proposed framework was subsequently evaluated on AMIGOS and EmoWear to assess cross-dataset and cross-device generalisation. AMIGOS presented an additional challenge: with only 15 participants in the included subset and approximately 24 windows per participant, each LOPO-CV fold trains on fewer than 336 examples. This lower number of windows relative to DREAMER's approximately 70 windows per participant  likely contributed to the lower cross-participant accuracy of 28.82\% before equalisation. These results show that participant level data availability is an important factor influencing the performance. EmoWear, with 36 participants and a more balanced LALV distribution of 219 windows, produced the most consistent four-class results.

\subsection{Binary Valence Classification and Algorithm Comparison}
Following the evaluation of four-class emotion classification, binary valence classification was investigated to determine whether reducing the classification task to positive and negative valence improves subject-independent performance. Binary valence classification collapses the four quadrants into HIGH (V above midpoint) and LOW (V at or below midpoint). This reduction is motivated by both practical and empirical considerations. From a practical prospective, distinguishing positive from negative emotional state is sufficient for mood monitoring. 
 Six classifiers were evaluated under identical conditions to determine the most suitable classifier for subject-independent binary valence classification. LR achieved the highest LOPO-CV accuracy 53.17\% while SVM  achieved the lowest 49.91\%. The summary of results is presented in Table~\ref{tab:algorithms}.
\vspace{-0.5em}
\begin{table}[htbp]
\caption{Algorithm Comparison: DREAMER Binary Valence LOPO-CV}
\label{tab:algorithms}
\centering
\small
\begin{tabular}{lccc}
\toprule
\textbf{Classifier} & \textbf{Accuracy} & \textbf{HIGH  recall} & \textbf{LOW recall} \\
\midrule
\textbf{Logistic Regression} & \textbf{53.17\%} & 41.4\% & 64.6\% \\
Random Forest & 52.45\% & 46.5\% & 58.2\% \\
LDA & 52.36\% & 41.0\% & 63.4\% \\
XGBoost & 51.90\% & 48.0\% & 55.7\% \\
KNN (k=5) & 51.45\% & 48.5\% & 54.3\% \\
SVM (RBF) & 49.91\% & 26.3\% & 72.9\% \\
\bottomrule
\end{tabular}
\end{table}
\vspace{-0.5em}

To evaluate the generalisability of binary valence classification across different datasets, the two best-performing classifiers (LR and RF) were subsequently evaluated on the DREAMER, AMIGOS, and EmoWear datasets. Table~\ref{tab:binary_all} summarises the corresponding LOPO-CV results. The results demonstrate consistent binary valence classification across the three datasets under the LOPO-CV protocol. LR achieved the highest accuracy on DREAMER (53.17\%) and AMIGOS (56.76\%), whereas RF performed best on EmoWear (55.51\%) and the combined dataset (53.37\%).

\vspace{-0.5em}
\begin{table}[htbp]
\caption{Binary Valence LOPO-CV Results Across Datasets}
\label{tab:binary_all}
\centering
\small
\begin{tabularx}{\columnwidth}{Xccc}
\toprule
\textbf{Dataset} & \textbf{RF} & \textbf{LR} & \textbf{Class Balance}  \\
\midrule
DREAMER & 52.45\% & 53.17\% & 544H / 560L \\
AMIGOS & 50.12\% & 56.76\% & Near-balanced \\
EmoWear & 55.51\% & 50.87\% & 632H / 456L \\
Combined & 53.37\% & N/A & All three \\
\bottomrule
\end{tabularx}
\end{table}
\vspace{-0.5em}

 Table~\ref{tab:comparison} compares the proposed framework with representative ECG-HRV emotion recognition studies.
The within-participant accuracy  on the DREAMER dataset reaches 65.13\%, exceeding the original SVM baseline of 62.49\%, using   multi-window HRV features framework. 
within-participant studies train and test on data from the same individuals,  which have generally produced higher accuracies than subject independent evaluations. Furthermore, the highest-accuracy prior work relies on raw ECG waveforms processed through deep learning on laboratory-grade equipment, or on person-specific feature extraction methods that are computationally unsuitable for edge deployment. 
\vspace{-0.5em}
\begin{table}[htbp]
\caption {Comparison of 
 ECG-HRV valence classification methods. 
 }
\label{tab:comparison}
\centering
\footnotesize
\setlength{\tabcolsep}{3pt}
\begin{tabularx}{\columnwidth}{Xcccc}
\toprule
\textbf{Study} & \textbf{Dataset} & \textbf{Method} & \textbf{Protocol} & \textbf{Accuracy} \\
\midrule
Katsigiannis~\cite{katsigiannis2018dreamer} & DREAMER & SVM & WP & 62.49\% \\
Nita et al.~\cite{nita2022cnn} & DREAMER & CNN & WP & 95.16\% \\
Wang et al.~\cite{wang2023hrv} & DREAMER & LR+ALQ & WP & 84.30\% \\
Ferdinando~\cite{ferdinando2016ecg} & Mahnob-HCI & SVM+EMD & XP & 42.60\% \\
\midrule
This study & DREAMER & RF & LOPO-CV & 52.45\% \\
This study & DREAMER & LR & LOPO-CV & {53.17\%} \\
This study & EmoWear & RF & LOPO-CV & 55.51\% \\
This study & DREAMER & RF & WP & 65.13\% \\
\bottomrule
\end{tabularx}
\end{table}
\vspace{-0.5em}
\section{Deployment and Real-World Validation}\label{sec:dep}

\subsection{Protocol for Live Validation }

To bridge the gap between dataset evaluation and real-world deployment, a live validation session was conducted using a Polar H10 chest strap  to record ECG signals at 130 Hz. 
The participant, one of the study authors, had no personal data included in the training dataset and received no calibration session, constituting a genuine zero-shot deployment scenario.

Four video clips were selected to elicit each of the four intended valence-arousal quadrants: a Bob Ross painting tutorial (LAHV), the opening montage of \textit{Up} (HAHV), a performance scene from \textit{Whiplash} (HALV), and a scene from \textit{Schindler's List} (LALV). After each clip, the participant self-reported valence and arousal on the DREAMER 1--5 scale. 

The recorded ECG was processed using the same preprocessing, feature extraction, classification and majority voting based final clip classification procedures  used during the training sessions.  
RF classifier was used across four-class valence classification and binary class classification on DREAMER dataset to evalute using both evalaution protocols including LOPO-CV and within the participant.
Self-reported ratings indicated that the participant's emotional responses did not always correspond to the intended valence-arousal quadrant. 

\begin{figure}[htbp]
\centerline{\includegraphics[width=\columnwidth]{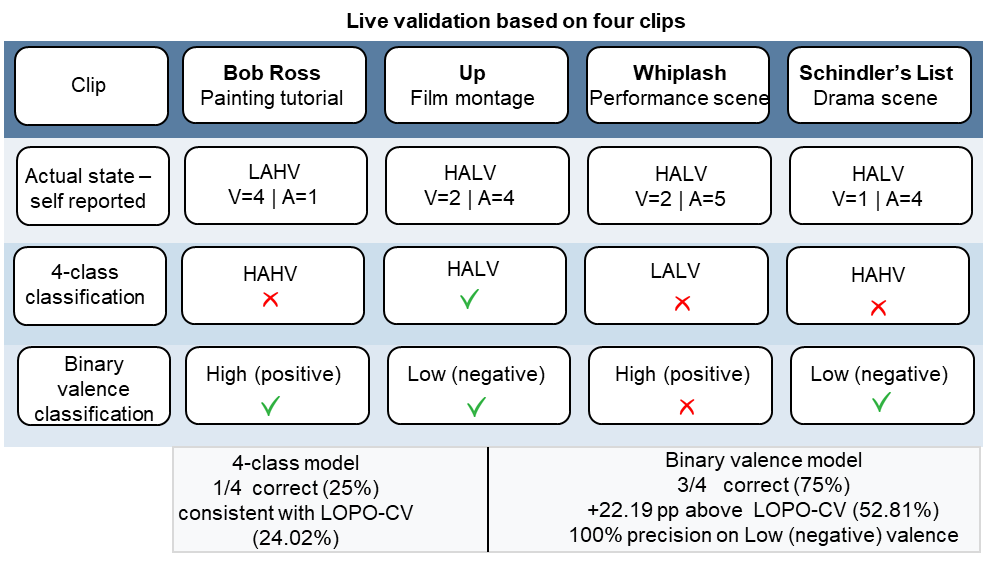}}
\caption{Live Polar H10 validation results for four video clips.
}
\label{fig:live}
\end{figure}

Fig.~\ref{fig:live} presents the  results using participant's self-reported emotional states as the ground truth. The four-class model achieved 25\%, which is consistent with the LOPO-CV result of 24.02\%. The binary valence model achieved 75\%, exceeding the equalised LOPO-CV result (52.81\%). 

\vspace{-0.5em}
\vspace{-0.5em}



\section{Proposed Method Performance in Federated Learning} \label{sec:per}







The federated system was implemented using the Flower framework \cite{beutel2020flower} and deployed on a physical cluster of six Raspberry Pi 4 devices. 
 One device (master server) runs the Flower server managing round coordination, weight aggregation, and global model storage. 
 Each of the remaining five devices holds a private data partition and returns weight updates to the server. The DREAMER dataset was partitioned by participant ID, with each node containing approximately four to five participants. No raw HRV features were shared between nodes at any point. The server received only model weight vectors of dimension $d = 9 \times C$, where $C$ is the number of output classes.
 To evaluate the effect of federated learning, the federated LR model is compared with its centralised LR counterpart under the same experimental setting.


ECG-HRV features are not independent and identically distributed) across participants \cite{frontiers2024federated}. Each individual has a unique cardiac autonomic profile that occupies a different region of the HRV feature space. Therefore, when a centralised LR model is trained on all 23 DREAMER participants simultaneously, it fits a single  decision boundary across all overlapping physiological distributions, producing a compromise suboptimal for each individual.

\begin{figure}[htbp]
\centerline{\includegraphics[width=\columnwidth,height=0.5\textheight,keepaspectratio]{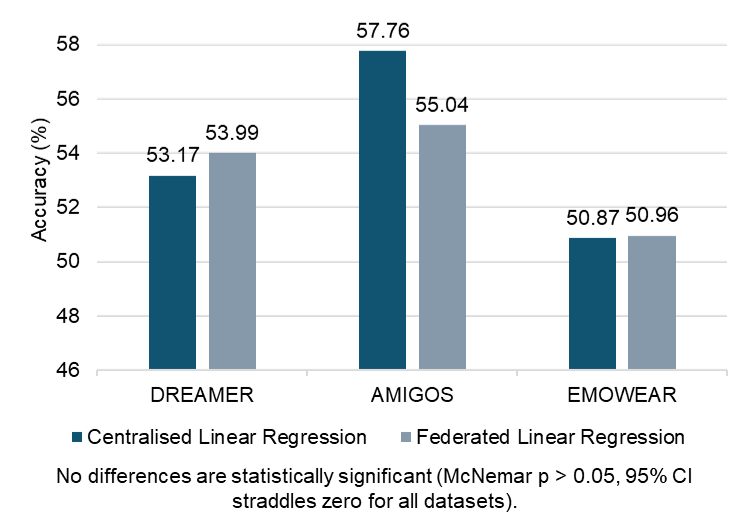}}
\caption{Centralised vs. federated Logistic Regression performance.}
\label{fig:convergence}
\end{figure}

In contrast, each federated node learns a locally calibrated decision boundary from its assigned participant data.
 FedAvg averages these into a global model informed by five independent optima rather than a single joint fit, with the averaging also acting as an implicit regulariser. As the system matures and each node accumulates data from a single user, local models will increasingly reflect that individual's physiological profile before contributing to the global update. 
Fig.~\ref{fig:convergence} compares the classification accuracy of centralised and federated LR across the DREAMER, AMIGOS, and EmoWear datasets. Federated learning achieved a modest improvement on the DREAMER dataset (53.99\% vs. 53.17\%), while centralised learning performed better on AMIGOS (57.76\% vs. 55.04\%) and both approaches produced comparable performance on EmoWear (50.87\% vs. 50.96\%).
\vspace{-0.5em}
\vspace{-0.5em}

\section{Discussion} \label{sec:dis}

This study demonstrates a  proof of concept for continuous, private mood monitoring on consumer hardware, evaluated end-to-end from raw ECG through a physical federated cluster. Starting from the ECG of a completely unseen individual recorded on a chest strap,  the approach achieves 75\% valence classification accuracy with zero calibration, exceeds the cross-participant benchmark by 22 percentage points, and   100\% precision for negative valence classification.
The algorithm comparison shows that all six classifiers converge to the same narrow accuracy range between 49.9\% and 53.2\%, which reframes the research direction for cross-participant HRV emotion classification. The cross-participant performance ceiling appears to be  physiological rather than  algorithmic. This suggests that  personalisation, which the federated architecture enables as personal data accumulates, is a more promising research direction.


The live validation result, while limited to a single participant and four stimuli, provides evidence that dataset-only benchmarks cannot offer: the system works on real physiology, in real time, with real consumer hardware. The 22  \% improvement in the live result over the LOPO-CV benchmark suggests that within-session physiological coherence provides a natural form of implicit personalisation even without explicit calibration.
The proposed system shows significant clinical and social potential, reaching far beyond laboratory evaluation and academic contexts.
A clinician monitoring a patient with depression could receive a continuous longitudinal record of valence trends rather than relying on fortnightly self-report. A parent noticing withdrawal in a teenager could have an objective early signal without intruding on the teenager's daily experience. Individuals managing their own mental health could receive early warnings of mood deterioration before a relapse takes hold. In each of these scenarios, the critical properties are not accuracy in isolation but the combination of deployability on consumer hardware, interpretable features a clinician can reason about, and a privacy model that keeps raw physiological data on the user's own device at all times. 

\subsection{Limitations and Ethical Considerations}
Despite these promising results, the proposed approach has so far been validated on only a single participant. Future work will involve conducting large-scale live validation. This will require obtaining ethical approval from  institutional ethics committee, alongside informed consent from participants, ensuring they are fully aware of  the data collected and their right to withdraw at any stage.  Therefore, this proposed
 system is designed as a personal awareness tool, not a clinical diagnostic instrument. 
The current study has several limitations. The live validation involved a single participant and four stimuli. The federated implementation used  dataset partitions rather than  longitudinal personal data from individual users. 
The current deployment assumes a shared local network. A geographically distributed deployment would require encrypted transmission over HTTPS, but this does not compromise the privacy  since only model weights  are transmitted, not raw signals \cite{mcmahan2017fedavg}.

\section{Conclusion} \label{sec:conc}

This paper  presented a systematic, multi-phase investigation of ECG-HRV based emotion classification for mood detection, from dataset evaluation to live wearable deployment to privacy-preserving federated learning. The key contributions are: the identification of class imbalance as the dominant suppressor of depressive mood state detection in four-class valence-arousal problems; the demonstration that standard nine-feature HRV classification matches 78-feature extended sets under cross-participant evaluation while remaining clinically interpretable; the first published standalone ECG-HRV binary valence classification results on the EmoWear dataset; live pilot validation on an unseen participant achieving 75\% binary valence accuracy 
on consumer hardware, a promising result pending broader participant validation; and federated learning achieving accuracy statistically equivalent to centralised training across three independent datasets on a Raspberry Pi cluster, with a  privacy preservation support and no sharing of raw physiological data. Future work will extend the federated architecture to incorporate EEG signals alongside the ECG approach, and will include live validation on a broader participant cohort, enabling a multimodal and generalised federated mood detection system. 

\section*{Acknowledgements}

For the purpose of open access, the author has applied a Creative Commons Attribution (CC BY) licence to any Author Accepted Manuscript version of this paper arising from this submission.

\bibliography{references}
\bibliographystyle{IEEEtran}

\end{document}